\documentclass[conference]{IEEEtran}
\IEEEoverridecommandlockouts
\usepackage{cite}
\usepackage{amsmath,amssymb,amsfonts}
\usepackage{algorithmic}
\usepackage{graphicx}
\usepackage{textcomp}
\usepackage{xcolor}
\usepackage{graphicx}
\usepackage[caption=false]{subfig}
\usepackage{lipsum}
\usepackage{xspace}
\usepackage{hyperref}
\usepackage[first=1, last=100]{lcg}
\usepackage{flushend}

\def\BibTeX{{\rm B\kern-.05em{\sc i\kern-.025em b}\kern-.08em
    T\kern-.1667em\lower.7ex\hbox{E}\kern-.125emX}}
\begin{document}

\title{Tiny Robot Learning: Challenges and Directions for Machine Learning in Resource-Constrained Robots
\thanks{This material is based upon work supported by the National Science Foundation under Grant 2030859.
Any opinions, findings, conclusions, or recommendations expressed in this material are those of the authors and do not necessarily reflect those of the funding organization.}
}
\author{\IEEEauthorblockN{
Sabrina M. Neuman$^{1}$,
Brian Plancher$^{1}$,
Bardienus P. Duisterhof$^{2}$,
Srivatsan Krishnan$^{1}$,
Colby Banbury$^{1}$,\\
Mark Mazumder$^{1}$,
Shvetank Prakash$^{1}$,
Jason Jabbour$^{3}$,
Aleksandra Faust$^{4}$,\\
Guido C.H.E. de Croon$^{5}$,
and Vijay Janapa Reddi$^{1}$
}
\IEEEauthorblockA{
\textit{Harvard University$^{1}$, CMU$^{2}$, University of Virginia$^{3}$, Google Brain$^{4}$, Delft University of Technology$^{5}$}\\
\{sneuman@seas, brian\_plancher@g, srivatsan@g, cbanbury@g, markmazumder@g, sprakash@g, vj@eecs\}.harvard.edu,\\ bduister@andrew.cmu.edu, jjj4se@virginia.edu, aleksandra.faust@gmail.com, g.c.h.e.decroon@tudelft.nl
}
}

\maketitle

\begin{abstract}
Machine learning (ML) has become a pervasive tool across computing systems.
An emerging application that stress-tests the challenges of ML system design is \emph{tiny robot learning}, the deployment of ML on resource-constrained low-cost autonomous robots.
Tiny robot learning lies at the intersection of embedded systems, robotics, and ML, compounding the challenges of these domains.
Tiny robot learning is subject to challenges from size, weight, area,
and power (SWAP) constraints; sensor, actuator, and compute
hardware limitations; end-to-end system tradeoffs; and a large
diversity of possible deployment scenarios.
Tiny robot learning requires ML models to be designed with these challenges in mind, providing a crucible that reveals the necessity of holistic ML system design and automated end-to-end design tools for agile development.
This paper gives a brief survey of the tiny robot learning space,
elaborates on key challenges,
and proposes promising opportunities for future work in ML system design.

\end{abstract}


\section{Introduction}

Machine learning (ML) has become a pervasive technology,
and as it spreads beyond traditional computing platforms (e.g., servers and desktops) towards devices on the edge
(e.g., mobile, embedded, IoT, AR/VR, robotics, and other cyber-physical systems),
new design pressures and constraints arise that fundamentally impact the ML system design process.

An emerging application that stress-tests the challenges of designing ML for edge devices is \emph{tiny robot learning}, the deployment of ML on resource-constrained low-cost autonomous robots. These robots are lightweight (e.g., less than a pound, or under $\sim500$g) and can operate in small spaces,
making them a promising solution for applications ranging from emergency search and rescue~\cite{doi:10.1126/scirobotics.aaw9710,duisterhof2021sniffy,duisterhof2021tiny},
to routine monitoring and maintenance of infrastructure and equipment~\cite{doi:10.1126/scirobotics.aau3038}.

\begin{figure}[t]
    \centering
    \includegraphics[width=1.00\columnwidth]{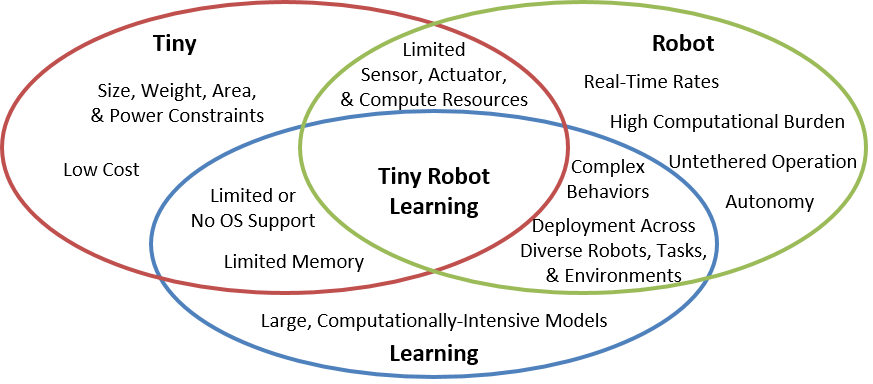}
    \caption{Tiny robot learning combines tiny (i.e., embedded) systems, robotics, and machine learning, revealing challenges and opportunities for ML research.}
    \label{fig:venn_diagram}
    \vspace{-1em}
\end{figure}

Tiny robot learning dials up the challenges of edge device ML,
maximizing opportunities to refine edge ML system design by putting it through the crucible of the combined challenges of tiny (i.e., embedded) systems, robotics, and machine learning, all in one system deployment (Fig.~\ref{fig:venn_diagram}).
Tiny robot learning is subject to challenges from size, weight, area, power (SWAP) and cost constraints; sensor, actuator, and compute hardware limitations; end-to-end system tradeoffs; and a large diversity of possible deployment scenarios.

Embedded \emph{tiny} systems have severe SWAP constraints. For machine learning, this means that large, computationally-intensive models must be made to run on computing substrates, such as microcontrollers, with limited memory resources, and limited or no operating system support.
Outside of robotics, SWAP constraints appear in many emerging edge device applications, and are being addressed with novel ``TinyML'' techniques such as model compression and distillation~\cite{banbury2020benchmarking}.


This challenge is compounded when tiny systems and machine learning intersect with robotics applications, and
small-scale low-cost \emph{tiny robots} impose additional constraints on both the ML model design as well as the end-to-end system surrounding the ML core.
These robots have limited sensors, actuators, and compute resources, which complicate the development and implementation of TinyML models.
Robotics is also a computationally-demanding application space,
where ML models must learn complex and robust robot behaviors.


Adding to this difficulty,
ML in untethered
autonomous robots is fundamentally \emph{embodied},
that is, the entire end-to-end robot system impacts the choices made in ML system deployment.
In a tiny robot especially, critical tradeoffs must be made between the power and weight resources allocated to ML versus other parts of the system, e.g., sensor processing.

Finally, tiny robot deployments vary across robot models, system components, tasks, and environments,
so it is essential to develop automatable flows
to keep the design process agile.


The challenges of tiny robot learning are a call to action for the ML circuits, architecture, and systems design community.
There are exciting opportunities for ML revealed by these challenges, including applying embedded TinyML techniques to computationally burdensome problems in robotics; using ML to compensate for the limitations of low-cost sensors and actuators; performing end-to-end co-design of ML with the surrounding cyber-physical system; and creating generalizable and automatable design flows for both software and hardware.

To explore the implications of tiny robot learning, in this work we
give a brief survey of the tiny robots space,
elaborate on the challenges imposed by tiny robot learning as an application for ML system design, and
propose opportunities revealed by these challenges to improve ML system design.





\section{Tiny Robots}
\begin{figure}[t]
    \centering
    \includegraphics[width=0.90\columnwidth]{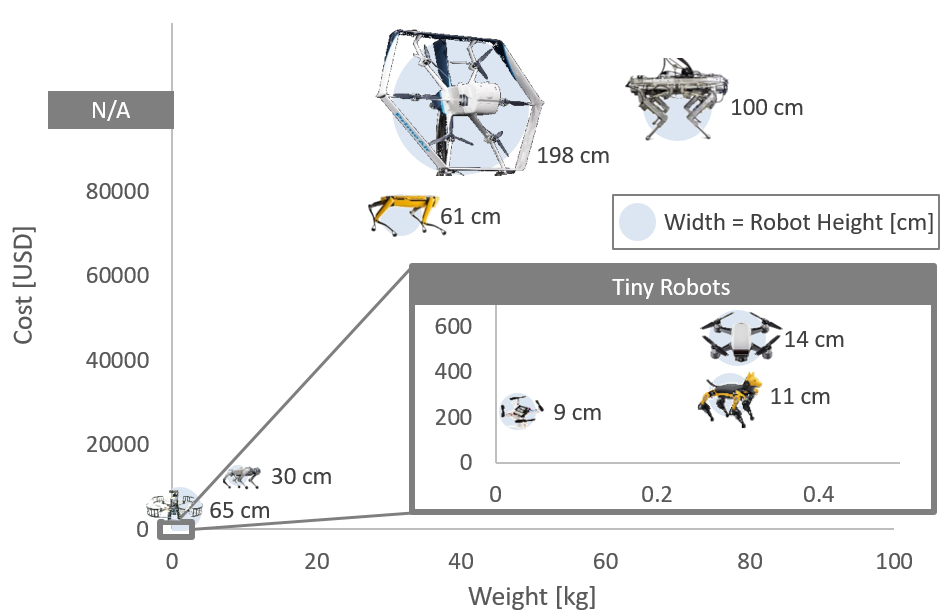}
    \caption{Tiny robots are resource-constrained low-cost autonomous robots. Pictured are  full-size~\cite{amazon2020MK27,bostondynamics2022spot,asctec2020pelican,semini2010design,katz2019mini} and tiny~\cite{petoi2022bittle,dji2022spark,bitcraze2022crazyflie} robots.}
    \label{fig:quadruped_scale}
\end{figure}

We define tiny robots as resource-constrained, low-cost, low-weight autonomous robots.  
An example 
is the Petoi Bittle quadruped~\cite{petoi2022bittle}, which weighs $0.64$~lbs ($290$~g) and costs under $300$~USD (Fig.~\ref{fig:quadruped_scale}). By contrast, the Boston Dynamics Spot quadruped~\cite{bostondynamics2022spot} weighs $69.9$~lbs ($31.7$~kg) and costs $74,500$~USD, making it $100\times$ heavier and more expensive than Bittle.
%
To fit within small weight and cost constraints, tiny robots are often limited in the availability and quality of onboard sensors, actuators, and compute resources, e.g., microcontrollers versus desktop-grade CPUs and GPUs.


While the same general challenges arise from ML design for a variety of different types of robots (e.g., quadrotor drones, satellites, quadrupeds, cars, submersibles), some design considerations differ between these platforms.
For example, when comparing quadrotors to quadrupeds, weight is a more extreme constraint for quadrotors, while quadrupeds require more computationally expensive motion planning and control algorithms due to their increased degrees of freedom.

Tiny robots are useful in a diverse range of applications.
Table~\ref{tab:applications_survey} shows a brief (and incomplete) overview of emerging applications of tiny robots ranging from search and rescue, to inspection, entertainment, and STEM education.
%
Because of their small size and ability to escape detection, we note that it is essential to exercise ethical consideration in the design and deployment of tiny robot systems~\cite{luppicini2016technoethical,arrieta2020explainable}. Concerns such as safety, privacy, and security must be factored into the engineering process as first-class constraints.

The tiny robot examples surveyed in Table~\ref{tab:applications_survey} demonstrate the effectiveness of co-design between sensors, compute and algorithms. These robots use sensors beyond traditional cameras: tiny lasers, optic flow sensors, light sensors, gas sensors, pressure sensors, and custom-made tiny cameras. The algorithms they use were intentionally designed with the compute constraints in mind, allowing them to run on low-power and low-cost microcontrollers. We expect novel sensors such as event cameras and optic flow sensors, as well as novel compute platforms like the Intel Lohi neuromorphic chip~\cite{davies2018loihi}, to greatly improve the performance and capabilities of future tiny robots.

\begin{table}[!t]
   \footnotesize
   \renewcommand{\arraystretch}{1.3}
   \caption{Emerging Tiny Robot Applications}
   \label{tab:applications_survey}
   \centering
   \resizebox{\linewidth}{!}{%
   \begin{tabular}{|l|l|l|l|}
      \hline
      Task          & Robot     & Weight [g]  & Citation   \\ \hline\hline
      Search \& Rescue          & Bitcraze CrazyFlie Quadcopter     & 33    & \cite{doi:10.1126/scirobotics.aaw9710,duisterhof2021sniffy,duisterhof2021tiny}   \\ \hline
      Inspection    & HAMR-E     & 1.4   & \cite{doi:10.1126/scirobotics.aau3038}   \\ \hline
      Medical Robotics          & Wireless Capsule Endoscope     & 7   &  \cite{endoscopy_bot}  \\ \hline
      Space Robotics          & KickSat     & 5   &  \cite{kicksat}  \\ \hline
      Military Reconnaissance          & Black Hornet     & 33  &  \cite{doi:10.1142/S2301385014300017}  \\ \hline
      Entertainment          & DelFly Nimble     & 28  &  \cite{doi:10.1126/science.aat0350,doi:10.1142/S2301385020500235}  \\ \hline
      STEM Education  & Mona     & 290   &  \cite{arvin2019mona}  \\ \hline
      Pollination  & RoboBee     & 0.08   &  \cite{wood2013flight}  \\ \hline
   \end{tabular}
   }
\end{table}

\section{Challenges and Opportunities}
Tiny robot learning lies at the intersection of three challenging domains (Fig.~\ref{fig:venn_diagram}), making it a proving-ground for ML systems.
In this section, we examine four themes arising from the challenges of tiny robot learning, and propose opportunities that they reveal for improving ML system design (Fig.~\ref{fig:opportunities}).

Starting from the ML computation at the core of the system and moving outward, we focus on: (\ref{subsec:swap-constrained-compute}) the onboard compute; (\ref{subsec:limited-sensors-actuators}) the sensors and actuation that represent the inputs and outputs of the robotics computation pipeline; (\ref{subsec:end-to-end-co-design}) the entire end-to-end system including the physical robot platform; and finally, (\ref{subsec:automated-design-flows}) the design tools used to re-design the tiny robot learning system for different deployment scenarios.

\subsection{SWAP-Constrained ML Compute for Robotics Applications}
\label{subsec:swap-constrained-compute}
In untethered robot systems, 
a subset of core computations must always be performed using onboard compute resources (rather than being offloaded to the cloud) 
due to the strict latency requirements of running at real-time rates (1kHz or more), 
as well as a lack of continuous communication and connectivity guarantees. 
Tiny robots have limited compute resources (e.g., microcontrollers) that are severely size, weight, area, and power (SWAP) constrained,
yet they must handle the same heavy computational burdens and produce the same complex robust behaviors in the real world as full-size autonomous robots with access to onboard server-class CPUs and GPUs.

Outside of robotics, SWAP constraints appear in many emerging edge computing applications (e.g., IoT), and are being addressed with TinyML techniques such as quantization~\cite{han2015deep}, distillation~\cite{hinton2015distilling}, and tiny model design~\cite{banbury2021micronets}.

\begin{figure}[!t]
    \centering
    \includegraphics[width=0.99\columnwidth]{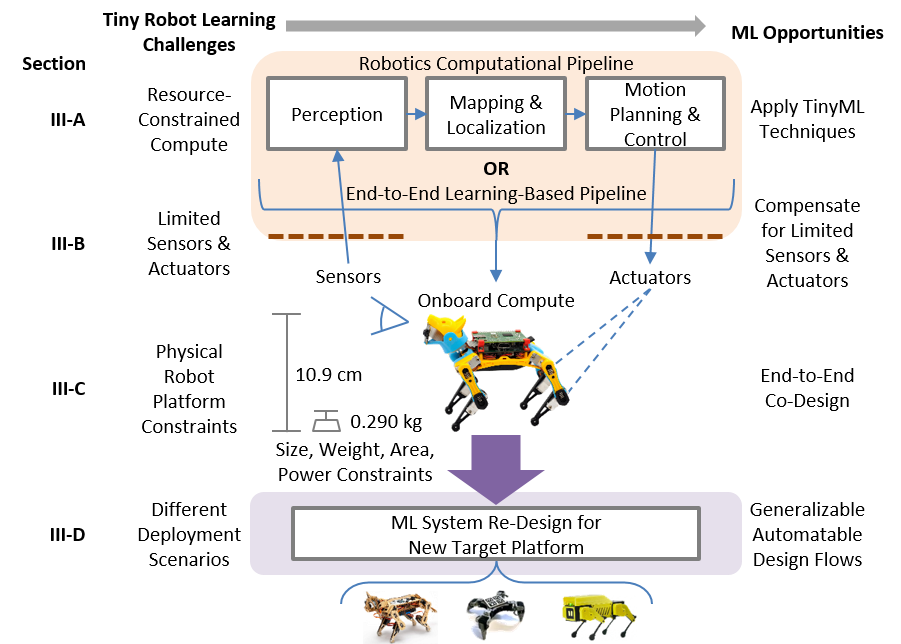}
    \caption{Promising research directions revealed by tiny robot learning include applying embedded \emph{TinyML} techniques to traditional robotics applications; using ML to compensate for sensor and actuator limitations; end-to-end co-design of ML with physical systems; and developing automatable design flows for agile re-deployment to new target platforms (e.g.,\cite{petoi2022bittle,stanford2022pupper,stanford2022realant}).}
    \label{fig:opportunities}
\end{figure}

\emph{Opportunity:}
Tiny robot learning reveals opportunities for the application of TinyML techniques to difficult problems in robotics.
Prior work has applied ML across the stages of the robotics computational pipeline (shown in Fig.~\ref{fig:opportunities}): perception~\cite{mahjourian2018unsupervised}, mapping and localization~\cite{liu2018semi}, and motion planning and control~\cite{kaufmann2018deep, lee2020learning}.
By mapping these robotics tasks to ML, one can leverage existing work to compress and accelerate ML under extreme SWAP constraints, enabling the deployment of complex tasks on resource-constrained tiny robot platforms.
Learned approaches can also offer computationally cheaper alternatives to traditional robotics algorithm pipelines through end-to-end learning-based approaches (Fig.~\ref{fig:opportunities})~\cite{duisterhof2021tiny,loquercio2018dronet}.

In addition to inference model implementation, there are implications for training in robot learning that take on a new dimension when applied to tiny robots. The mainstream approach is to learn offboard and deploy the trained solution on the robot. An obvious limitation of small robots is that the solution, typically a neural network model, has a lower complexity (fewer layers, neurons, etc.). Besides a somewhat lower performance, this also means that it is hard to train models on a large variety of environments and conditions, as the network may simply not be expressive enough.

A promising solution to this is to perform online learning, which would allow the tiny robot to learn about its immediate environment. Having a small learned model for a lower variety of environments may be acceptable, given that smaller robots will typically travel less far~\cite{7759284}. It would be interesting to investigate the trade-off between model complexity and practical use for tiny autonomous robots. Like small insects, tiny robots may be able to exploit less powerful learning models to function well in a specific environment~\cite{ERNST19993920}. Of course, online learning is already a challenge for robots in general and becomes even more challenging on tiny robots. For example, catastrophic forgetting is often solved by means of maintaining a replay buffer~\cite{mnih2015human}, but this will be harder to execute given very limited memory. In general, machine learning methods for tiny robots will have to more carefully assign resources to what is learned and what is forgotten.

\subsection{Sensor and Actuator Limitations in Tiny Robot Platforms}
\label{subsec:limited-sensors-actuators}

With cost and SWAP constraints, traditional sensors are often impractical on tiny robot platforms, forcing them to rely on lower-quality sensors~\cite{duisterhof2021tiny}.
Tiny robots may also have fewer and less-precise actuators than full-size robots. For example, the Spot quadruped~\cite{bostondynamics2022spot} has position and force sensors on its legs and 12 degrees of freedom (DoF), while Bittle~\cite{petoi2022bittle} has no position or force sensors and only 8 DoF. 
ML techniques can enable complex behaviors on full-sized robots 
(e.g., quadruped locomotion on difficult terrain~\cite{lee2020learning}), 
but with component limitations, tiny robot platforms face an additional hurdle.


\emph{Opportunity:}
Low cost proximity and light sensors can replace larger, more costly sensors like cameras or LIDAR by using additional processing, often ML, to achieve the same levels of perception as their higher-quality counterparts.
For example, monocular depth estimation~\cite{mahjourian2018unsupervised} can be substituted for wide-baseline stereo cameras. In some cases, the lower dimensional input from, e.g., light or proximity sensors,
means that a simple policy can be developed to solve a complex problem.
For example, system-specific sensor and algorithm selection can yield lighter computation than traditional mapping and localization using cameras or LIDAR~\cite{duisterhof2021tiny}.
ML algorithms can also enable robots to use limited and degraded sensory input to
overcome challenges such as motion control with imprecise actuators and a lack of direct position feedback.
For example, recent work used a hybrid model-based and learning-based controller to enable a tiny quadruped to walk over uneven terrain~\cite{rahme2021linear}, despite tight SWAP and cost constraints on sensing and actuation.


\subsection{End-to-End Co-Design of ML with Physical Robot Systems}
\label{subsec:end-to-end-co-design}
Given the severe software and hardware constraints faced by tiny robots (Sections~\ref{subsec:swap-constrained-compute},~\ref{subsec:limited-sensors-actuators}) there is a need for holistic end-to-end system co-design to develop optimized and task-specific tiny robot systems.
Designers must account for physical properties (e.g., battery weight) to understand overall system metrics like maximum velocity~\cite{krishnan2020sky} and mission time.

\emph{Opportunity:}
There is an opportunity to improve overall system performance by building holistic end-to-end benchmarking frameworks.
These frameworks enable exploration of quantitative tradeoffs between cost, robustness, and efficiency across the entire cyber-physical stack, and help designers find efficient operating points, which are especially important for resource-constrained platforms.
An example is the Air Learning~\cite{krishnan2021air} framework, which evaluates combinations of learning algorithms, ML models, sensing modalities, and onboard compute platforms for autonomous drones. 

By leveraging end-to-end benchmarks, the physical properties of robotic systems can be co-designed with the computer hardware, algorithms, and application.
Early work has explored co-design for drones~\cite{carlone2019robot,krishnan2020sky} and legged robots~\cite{fadini2021computational}.

An end-to-end benchmarking framework also exposes design parameters that can enable the use of machine learning methods \emph{to build machine learning systems}. Recent work has used techniques like Bayesian optimization~\cite{reagen2017case}, evolutionary algorithms~\cite{chang2021hardware}, and reinforcement learning~\cite{mirhoseini2020placement} to efficiently navigate the design space of neural network model parameters. These techniques can be extended to co-design ML hardware accelerators with the physical parameters of the robot platform.

\subsection{Generalizeable and Automatable Design Flows}
\label{subsec:automated-design-flows}
An extreme challenge for tiny robot learning is navigating the large diversity of robot deployment scenarios.
Deployments vary across robots, system components, tasks, and environments, creating an enormous design space to explore.


\emph{Opportunity:}
The diversity of robot learning deployments reveals the need for generalizeable and automatable design flows and methodologies for efficiently navigating the large design space.
There is emerging work in designing such tools for non-learning-based robotics hardware accelerators~\cite{neuman2021robomorphic,plancher2021accelerating,liu2021archytas,sacks2018robox}.
%
Outside of the robotics domain, there is also early work on full-stack open-source tools and generalizable design flows for the rapid deployment of TinyML accelerators~\cite{prakash2022cfu}.

Some combination of the aforementioned works, along with the holistic benchmarking frameworks from Section~\ref{subsec:end-to-end-co-design}, may be useful in designing tools for tiny robot learning systems.
Automated design flows are especially appealing for tiny robot learning because they can encode domain-specific knowledge into the tools and allow for end-to-end system design without intervention from experts in multiple disparate domains: embedded systems, robotics, and ML (Fig.~\ref{fig:venn_diagram}).

\section{Conclusion}
In this work, we examined tiny robot learning: 
the deployment of ML on resource-constrained low-cost autonomous robots.
Lying at the intersection of embedded
systems, robotics, and ML, tiny robot learning is subject to challenges from size,
weight, area, and power constraints; sensor, actuator,
and compute hardware limitations; end-to-end system tradeoffs;
and a large diversity of possible deployment scenarios.
As such, it reveals promising opportunities for future work developing holistic ML system design techniques and automated end-to-end design tools for agile development.

\bibliographystyle{IEEEtranS}
\bibliography{references}

\end{document}